\documentclass[journal]{IEEEtran}
\usepackage{cite}
\usepackage{amsmath,amssymb,amsfonts}
\usepackage{algorithm}
\usepackage{algorithmic}
\usepackage{graphicx}
\usepackage{textcomp}
\usepackage{xcolor}
\usepackage{bm}
\usepackage{empheq}
\usepackage{booktabs,tabularx}
\newcolumntype{Y}{>{\raggedright\arraybackslash}X}

\newtheorem{remark}{Remark}
\newcommand{\vect}[1]{\boldsymbol{#1}} 
\newcommand{\mat}[1]{\boldsymbol{#1}}  
\newcommand{\diag}{\text{diag}}
\newcommand{\conj}[1]{\overline{#1}}   

\DeclareMathOperator*{\st}{s.t.}

\PassOptionsToPackage{hyphens}{url}\usepackage{hyperref}

\usepackage{setspace}

\begin{document}

\setlength{\abovedisplayskip}{1pt}
\setlength{\belowdisplayskip}{1pt}
\addtolength{\jot}{-0.2em}

\title{GradMAP: Gradient-Based Multi-Agent Proximal Learning for Grid-Edge Flexibility}


\author{ 
    \IEEEauthorblockN{Yihong Zhou*, Hongtai Zeng, and Thomas Morstyn}\\
    \IEEEauthorblockA{
    Department of Engineering Science, University of Oxford\\
    *\textit{Correspondence: }\href{mailto:yihong.zhou@eng.ox.ac.uk}{yihong.zhou@eng.ox.ac.uk}
    \vspace{-3 mm}
} 
   
}


\maketitle

\begin{abstract}
Coordinating large populations of grid-edge devices requires learning methods that remain decentralised in deployment while still respecting three-phase AC distribution-network constraints. This paper proposes gradient-based multi-agent proximal learning (GradMAP) to address this challenge. GradMAP trains independent neural network policies for each agent that makes online decisions without communication. During offline training, GradMAP embeds a differentiable power grid environment and uses implicit differentiation to propagate exact AC power flow gradients. To speed up training while maintaining performance, GradMAP \emph{reuses} expensive environment gradients through a proximal surrogate within a trust region defined in the more direct policy-output space, instead of the probability distribution space used in PPO.
{\color{black} Using a single workstation-class NVIDIA RTX PRO 5000 Blackwell GPU with 48~GB of memory, GradMAP trains 1{,}000 agents on the IEEE123 network in 15 minutes and 100{,}000 agents on the IEEE8500 network in 40 minutes, without parameter sharing. In both cases, GradMAP achieve the lowest mean daily maximum voltage violation and the lowest energy cost among all evaluated methods, including standard multi-agent reinforcement learning methods and gradient-based learning methods extended to the multi-agent setting by us. The training time to convergence is also the lowest, with a 5$\times$ to more than $10\times$ training speedup compared to other methods.}
\end{abstract}


\IEEEpeerreviewmaketitle

\vspace{-3mm}
\section{Introduction}
\vspace{-1mm}

Decarbonisation and the electrification of heating and transportation are driving rapid growth in the number of grid-edge resources, including distributed renewables, battery storage, electric vehicles, and heat pumps. These devices can provide valuable flexibility through controllable power consumption or injection. Collectively, they can offer flexibility on a scale comparable to multiple large conventional power plants, with their global value estimated at USD~\$270~billion by 2040 \cite{iea_value}. However, traditional centralised dispatch approaches become increasingly impractical as the number of devices grows because of both computational limits (large-scale optimisation under uncertainty) and communication constraints (latency and bandwidth across the grid edge).

To address these challenges, existing work has explored two broad classes of methods: traditional model-based coordination methods and, more recently, learning-based methods. Among traditional approaches, distributed optimisation solves the system-wide coordination problem through iterative message passing \cite{distributed_robust_ed, R1D}. Another traditional approach is hierarchical aggregation--disaggregation \cite{zhou2024aggregated, Agg_DSO_field_test}, in which a system operator can perform smaller-scale dispatch over aggregated feasible power regions submitted by aggregators, who then map aggregate instructions back to individual grid-edge devices. However, distributed optimisation often suffers from significant communication overheads, slow convergence, and numerical instability \cite{agg_partI}. Meanwhile, aggregation-based methods rely on accurate characterisations of the cost and feasible region of the aggregated flexibility resources, which is challenging due to uncertainty, nonlinear device dynamics, and nonlinear alternating-current (AC) distribution-network constraints \cite{zhou2024aggregated}.

Another promising direction is learning-based decision-making, including imitation learning (IL), reinforcement learning (RL), and self-supervised learning. In IL, an agent learns a policy by mimicking a set of expert decision trajectories, which may be obtained from historical operational data or generated by solving offline optimisation problems. This approach has been shown to be effective for power system dispatch \cite{zhou2021lstm, gao2021online}. However, it requires a significant amount of expert data, which is unlikely to be available for new systems and systems undergoing rapid change. Although expert training data can be generated through offline simulations, this is costly when the underlying decision problem is difficult to solve, such as for large-scale AC optimal power flow (OPF).

In RL, agents learn through interaction with an environment that can largely be treated as a black box, as long as observations and rewards are returned in response to actions. This broad applicability makes RL appealing for many decision-making problems \cite{sutton2018reinforcement, cao2020deep, qiu2023reinforcement}. Multi-agent (MA) RL (MARL) is a particularly relevant extension for coordinating large numbers of grid-edge devices, because single-agent RL would require an impractical \textcolor{black}{level of communication to monitor millions of devices and disseminate online dispatch signals to them. Centralised single-agent RL would also raise issues related to the autonomy and privacy of individual device owners. The MA setting addresses these issues by assigning each device (or group of devices) its own agent,} which can make online decisions without communication while learning coordination through offline training. Recent studies have explored MARL for grid-edge device coordination \cite{charbonnier2022scalable, qiu2023reinforcement, charbonnier2025centralised}. However, these works follow the standard black-box RL setting, treating the power system environment as a black box and neglecting the fact that power system operators have been using well-established grid physics and mathematical models for decades \cite{neso2025network, kaya2025fifty, chiang2023dynamic}. Ignoring available network modelling methods is unlikely to be information-efficient.

Some RL studies have also recognised the benefits of environment models, especially when they are differentiable. Refs. \cite{clavera2020model, liu2024deep, ye2025physics} showed that exploiting reward gradients w.r.t. actions can improve the efficiency of RL learning and lead to better constraint satisfaction. \textcolor{black}{A challenge here} is that evaluating such environment gradients can be computationally expensive. For example, AC power-flow gradients w.r.t. power injections typically need to be obtained through implicit differentiation. 
{\color{black}
To alleviate this cost, PODS \cite{mora2021pods} and GI-PPO \cite{son2023gradient} first used environment gradients to construct improved actions (or an action policy) for \emph{multi-step} supervised policy learning that improves efficiency. Ref.~\cite{zhong2025reparameterization} showed that such methods can lead to misaligned learning objectives that limit learning performance, and proposed RPO for direct \emph{environment gradient reuse} with probability-based trust-region clipping and a KL penalty to improve performance. This approach is similar to the well-known proximal policy optimisation (PPO) approach used in standard black-box RL \cite{ppo2017proximal}. However, it remains questionable whether KL- and clipping-based trust-region regularisation defined in the probability distribution space is still the best choice for the differentiable case, because a direct trust region in policy-output space is more closely aligned with the coordinates in which the gradients are defined. Moreover, these works \cite{mora2021pods, son2023gradient, zhong2025reparameterization} focused only on single-agent settings and non-power-grid problems.
}

Self-supervised learning trains models on unlabeled data by generating its own supervisory signals. In power system decision-making, existing works \cite{park2024self, chen2025neural, anrrango2026self} start from standard power system optimisation that finds optimal power dispatch, and then extend it to a more nonlinear non-convex problem that finds the optimal parameterised decision policies generating the dispatch, thereby removing the MDP assumption in RL. In this context, (environment) gradients are the commonly exploited self-supervisory signals. However, most existing methods require gradient evaluation at every policy update, leading to the same challenge of computationally expensive gradient evaluation previously noted for differentiable RL \cite{clavera2020model, liu2024deep, ye2025physics}. 
{\color{black} Ref.~\cite{chen2026hard} implemented a strategy similar to that of PODS \cite{mora2021pods}, using cached improved actions for multi-step supervised learning, so it faces a similar misalignment issue. Ref.~\cite{chen2024model} tried to train a surrogate environment model to accelerate gradient evaluation, but such a data-driven power-grid surrogate may introduce extra bias. Existing self-supervised learning literature \cite{park2024self, chen2025neural, chen2026hard, anrrango2026self, chen2024model} has also failed to consider the multi-agent setting.
}

As the above discussion shows, several key gaps remain in learning-based decision-making. For IL \cite{zhou2021lstm, gao2021online}, the main limitation is the difficulty of obtaining expert decision-making data for training. For RL, a major gap is that well-established power system models and their differentiability are often unexploited, leading to less efficient learning \cite{cao2020deep, qiu2023reinforcement, charbonnier2022scalable, charbonnier2025centralised}. \textcolor{black}{Although some RL studies and self-supervised-learning-based OPF methods have explored gradient information, they have either failed to consider more efficient gradient utilisation \cite{clavera2020model, liu2024deep, ye2025physics, park2024self, chen2025neural, anrrango2026self}, used supervised updates that can perform worse than direct gradient reuse \cite{mora2021pods, son2023gradient, chen2026hard}, or used direct gradient reuse with less effective regularisation \cite{zhong2025reparameterization}. Also, existing works have focused only on the single-agent setting, which requires substantial online inter-device communication when applied to coordinating a large number of grid-edge flexible devices.}

In this paper, we propose a \textit{gradient-based multi-agent proximal learning (GradMAP)} framework for large-scale grid-edge flexibility coordination that addresses all the identified gaps. 
\textcolor{black}{We adopt direct environment gradient reuse, propose more aligned and effective policy-output-space regularisation for proximal updates, and extend the approach to a multi-agent setting with local policies that require no online inter-device communication and are trained through exact differentiable three-phase AC power flow. Table~\ref{tab:response-related-work} summarises work relevant to GradMAP and our key distinction.}
\textcolor{black}{The main contributions of GradMAP include:}
\begin{enumerate}
    \color{black}
    \item \textcolor{black}{A gradient-based learning framework for grid-edge flexibility coordination. To alleviate the computational complexity in evaluating the environment gradient, we adopt direct environment gradient reuse and propose more aligned and effective policy-output-space regularisation for proximal updates.}
    \item \textcolor{black}{Extension to multi-agent learning that does not require online inter-device communication once the offline centralised training is completed.}
    \item \textcolor{black}{Exact differentiable three-phase unbalanced AC power flow in the training loop, allowing AC network feasibility to directly shape agent policy learning.}
    \item \textcolor{black}{Case studies show that the proposed GradMAP method can train 1,000 agents without parameter sharing within 15 minutes and 100,000 agents within 40 minutes on a single workstation GPU. Across the tested cases, GradMAP achieves the lowest or comparable operating cost and AC constraint violations. The training time to convergence of GradMAP is up to $133\times$ faster over MARL and $5.3\times$ over the multi-agent differentiable RL/self-supervised OPF reference without gradient reuse.}
\end{enumerate}
The paper is organised as follows. Section~\ref{sec:problem_formulation} presents the problem formulation. Section~\ref{sec:gradma} describes our proposed ``GradMA'' framework, which extends single-agent self-supervised learning to the MA setting \textcolor{black}{without reusing the environment gradient}. Section~\ref{sec:gradmap} describes the full \emph{GradMAP} framework. Section~\ref{sec:case_study} reports the experimental results. Section~\ref{sec:discussion} provides discussions and Section~\ref{sec:conclusion} concludes the paper.

\begin{table}[t]
{\color{black}
\centering
\scriptsize
\caption{Representative related works versus our proposed GradMAP.}
\label{tab:response-related-work}
\setlength{\tabcolsep}{2pt}
\renewcommand{\arraystretch}{1.10}
\begin{tabularx}{\linewidth}{@{}>{\raggedright\arraybackslash}p{0.19\linewidth} Y >{\raggedright\arraybackslash}p{0.20\linewidth} >{\raggedright\arraybackslash}p{0.18\linewidth}@{}}
\toprule
Method & Gradient use and update & Policy and execution (when applied to our problem) & Targeted problem \\
\midrule
Gradient-based RL~\cite{clavera2020model,liu2024deep,ye2025physics} & Evaluating environment gradients for each policy update. & Single-agent; central policy requires online inter-device communication. & Robotics, storage bidding and OPF. \\
PODS / GI-PPO~\cite{mora2021pods,son2023gradient} & Environment gradients construct improved actions or an action policy for \emph{multi-step} supervised policy learning. Performance can be lower than with direct gradient reuse \cite{zhong2025reparameterization}. & Same as above. & Robotics, physics and traffic. \\
RPO~\cite{zhong2025reparameterization} & \emph{Direct environment gradient reuse} with probability-ratio clipping and probability-space KL regularisation. & Same as above. & Differentiable control. \\
Self-supervised learning for OPF~\cite{park2024self,chen2025neural,chen2026hard,anrrango2026self} & OPF gradient evaluation for each policy update or multi-step supervised
policy learning using improved actions \cite{chen2026hard}. Performance can be lower than with direct gradient reuse \cite{zhong2025reparameterization}. & Same as above. & DC/AC OPF and dispatch. \\
IPPO / MAPPO~\cite{yu2022surprising} & PPO updates use rewards and value estimates multiple times, without environment gradients. & Multi-agent; support local online execution without inter-device communication. & Cooperative games. \\
Our proposed GradMAP & \emph{Direct environment gradient reuse} with \emph{policy-output-space regularisation}. & Multi-agent; local online execution \emph{without inter-device communication}. & Grid-edge coordination with unbalanced three-phase AC power flow. \\
\bottomrule
\end{tabularx}
}
\end{table}


\vspace{-1mm}
\section{Problem Formulation}\label{sec:problem_formulation}

We consider a population of homes connected to a three-phase unbalanced distribution feeder. Each home is represented by one agent, and each agent controls one flexible device, such as a battery, a heat pump, or a controllable generator, together with exogenous uncontrollable rooftop PV and local demand. This one-agent-per-home, one-device-per-home assumption is representative of many residential flexibility settings and is adopted for ease of presentation. The proposed framework could handle homes with multiple controllable assets either by augmenting the local state-action space or by introducing multiple coordinated agents within the same home. The proposed framework could also be straightforwardly extended beyond residential flexibility to a wider set of grid-edge coordination problems (e.g. EV charging networks, commercial battery storage systems).

The objective is to learn decentralised policies that minimise the operation cost of all homes while satisfying both local device constraints and feeder-level physical limits. The potential user of this learning framework may be an aggregator or the distribution system operator (DSO). We assume that this user has access to the network model, and this is possible since this information sharing supports network constraint handling.
Note that our focus is on the communication-free decision-making framework rather than fair profit-sharing or market design. Although we use the term ``flexibility'', demand baselining methodologies are not the focus here, which mainly affect market settlement. With this in mind, we first introduce the device models and then the three-phase network model that couples them.

\vspace{-3mm}
\subsection{Grid-Edge Device Models}\label{sec:device_model}
\vspace{-1mm}

We consider three representative classes of grid-edge devices that capture important cross-time couplings encountered in grid-edge flexibility: batteries with nonlinear efficiency and energy coupling, heat pumps with thermal coupling, and controllable distributed generators with ramp-rate coupling. Given our home-level abstraction, we next describe the dynamics, constraints, and cost function of each controllable device type.

\subsubsection{Battery}
For a battery managed by agent $i$, its battery dynamics, constraints, and operating cost are summarised as:
\begin{subequations}\label{eq:bat_model}
    \begin{empheq}[left=\hspace{3em}\empheqlbrace]{align}
    \hat{E}_{i,t+1} &= E_{i,t}+\eta_{i}(\hat{P}_{i,t}^{\text{bat}})\, \hat{P}_{i,t}^{\text{bat}}\, \Delta t, \label{eq:bat_dyn} \\
    E_{i,t+1} &= \mathrm{clip}(\hat{E}_{i,t+1},\,0,\,E_i^{\max}), \label{eq:bat_true_soc} \\
    P_{i,t}^{\text{bat}} &= (E_{i,t+1}-E_{i,t}) / ({\eta_i(\hat{P}_{i,t}^{\text{bat}})\,\Delta t}), \label{eq:bat_clipping}
    \end{empheq}
    \begin{empheq}[left=\hspace{-3.5em}\empheqlbrace]{align}
    -P_{i,\max}^{\text{bat}} &\le \hat{P}_{i,t}^{\text{bat}} \le P_{i,\max}^{\text{bat}}, \label{eq:bat_power_bounds} \\
    0 &\le \hat{E}_{i,t+1} \le E_i^{\max}, \label{eq:bat_soc_bounds} \\
    E_{i,T} &\ge E_{i,\text{target}}^{\text{bat}}, \label{eq:bat_soc_end}
    \end{empheq}
    \vspace{-2mm}
    \begin{empheq}[left=\empheqlbrace]{align}
    C_{i,t}^{\text{bat}} &= C_{i,t}^{\text{deg}} + C_{i,t}^{\text{energy}}, \label{eq:bat_cost} \\
    C_{i,t}^{\text{deg}} &= c_i^{\text{deg}} |P_{i,t}^{\text{bat}}| \Delta t, \label{eq:bat_deg_cost} \\
    C_{i,t}^{\text{energy}} &= \rho_t^{\text{imp}} [p_{i,t}^{\text{net}}]^+ - \rho_t^{\text{exp}} [-p_{i,t}^{\text{net}}]^+, \label{eq:bat_energy_cost} \\
    p_{i,t}^{\text{net}} &= P_{i,t}^{\text{load}} - P_{i,t}^{\text{pv}} + P_{i,t}^{\text{bat}},
    \label{eq:bat_net}
    \end{empheq}
\end{subequations}
where $E_{i,t}$ denotes the stored energy, and $\hat{P}_{i,t}^{\text{bat}}$ denotes the power command issued by the agent (control variable). $t$ is the time step within the control horizon $\mathcal{T} := \{1, \dots, T\}$.
For interpretability, we define an auxiliary variable $\hat{E}_{i,t+1}$ to represent the \textit{implied next energy state} if the battery follows exactly the power command $\hat{P}_{i,t}^{\text{bat}}$ for the whole time step $\Delta t$ without any physical limits. However, in the real world, the battery stops charging or discharging once the stored energy reaches its bounds, which is captured by the clipping operation in \eqref{eq:bat_true_soc}. Therefore, the actual AC-side power $P_{i,t}^{\text{bat}}$, averaged over $\Delta t$, should be back-calculated from the true energy change and the efficiency $\eta_i(\hat{P}_{i,t}^{\text{bat}})$, as shown in \eqref{eq:bat_clipping}. The power command $\hat{P}_{i,t}^{\text{bat}}$ is constrained by \eqref{eq:bat_power_bounds}, and the implied next energy state $\hat{E}_{i,t+1}$ is constrained by \eqref{eq:bat_soc_bounds} (although mathematically redundant, its violation provides valuable learning signals as will be discussed in Section \ref{sec:exact_grad_pd}). We also introduce an end-state energy requirement in \eqref{eq:bat_soc_end}, which ensures the battery retains sufficient stored energy for the next episode.

Instead of making the common assumption of constant charging and discharging efficiency used in much of the battery optimisation literature (which provides computational tractability but can introduce a significant ``sim-to-real'' gap), we use a set of more accurate nonlinear efficiency curves obtained by exponential fitting to measurement data \cite{thingvad2017assessing}, which captures the reduced efficiency typically observed at low power levels. Our case studies randomly assign a curve to each battery system. 

Eqs.~\eqref{eq:bat_cost}--\eqref{eq:bat_net} separate throughput-based battery degradation from the electricity cost (with import/export tariffs $\rho_t^{\text{imp}}$ and $\rho_t^{\text{exp}}$) at the household meter. Note that in full detail, degradation cost is also nonlinear and consists of both calendar aging and cycle aging \cite{naumann2020analysis}. For simplicity we do not include these features, but our inclusion of nonlinear efficiency shows the framework could straightforwardly be extended to also incorporate nonlinear degradation cost.

\subsubsection{Heat Pump}
The model for a heat pump is:
\begin{subequations}\label{eq:hp_model}
    \begin{empheq}[left=\empheqlbrace]{align}
    \hat{\vartheta}_{i,t+1} &=
    \vartheta_{i,t} + \frac{\Delta t}{C_i}
    \Big(\frac{\vartheta_t^{\text{out}}-\vartheta_{i,t}}{R_i} + \mathrm{COP}_i \hat{P}_{i,t}^{\text{hp}}\Big), \label{eq:hp_dyn} \\
    \vartheta_{i,t+1} &= \mathrm{clip} \big(\hat{\vartheta}_{i,t+1},\,\vartheta_i^{\text{set}}-\delta_i,\,\vartheta_i^{\text{set}}+\delta_i\big), \label{eq:hp_true_temp}
    \end{empheq}
    \begin{empheq}[left=\hspace{-4em}\empheqlbrace]{align}
    P_{i,t}^{\text{hp}} &= \frac{\frac{C_i}{\Delta t} (\vartheta_{i,t+1}-\vartheta_{i,t} )-\frac{\vartheta_t^{\text{out}}-\vartheta_{i,t}}{R_i}}{\mathrm{COP}_i}, \label{eq:hp_actual_power} \\
    0 &\le \hat{P}_{i,t}^{\text{hp}} \le P_{i,\max}^{\text{hp}}, \label{eq:hp_power_bounds} \\
    \vartheta_i^{\text{set}}-\delta_i &\le \hat{\vartheta}_{i,t+1} \le \vartheta_i^{\text{set}}+\delta_i, \label{eq:hp_temp_bounds} \\
    \vartheta_{i,T} &\ge \vartheta_{i,\text{target}}^{\text{hp}}, \label{eq:hp_temp_end}
    \end{empheq}
    \vspace{-2mm}
    \begin{empheq}[left=\hspace{-5.75em}\empheqlbrace]{align}
    C_{i,t}^{\text{hp}} &= C_{i,t}^{\text{use}} + C_{i,t}^{\text{energy}}, \label{eq:hp_cost} \\
    C_{i,t}^{\text{use}} &= c_i^{\text{deg}} |P_{i,t}^{\text{hp}}| \Delta t, \label{eq:hp_use_cost} \\
    C_{i,t}^{\text{energy}} &= \rho_t^{\text{imp}} [p_{i,t}^{\text{net}}]^+ - \rho_t^{\text{exp}} [-p_{i,t}^{\text{net}}]^+, \label{eq:hp_energy_cost} \\
    p_{i,t}^{\text{net}} &= P_{i,t}^{\text{load}} - P_{i,t}^{\text{pv}} + P_{i,t}^{\text{hp}},
    \label{eq:hp_net}
    \end{empheq}
\end{subequations}
where $\vartheta_{i,t}$ denotes the indoor temperature, and $\hat{P}_{i,t}^{\text{hp}}$ denotes the electrical power command issued by the agent (control variable). Similar to the battery model, here we have defined an auxiliary variable $\hat{\vartheta}_{i,t+1}$ to represent the \textit{implied next indoor temperature} if the heat pump operates fully according to the power command $\hat{P}_{i,t}^{\text{hp}}$. Here we assume that the indoor temperature will also stop at the comfort boundaries through a feedback controller, which is captured by the clipping operation in \eqref{eq:hp_true_temp}. Therefore, the actual AC-side power $P_{i,t}^{\text{hp}}$ (averaged over $\Delta t$) should be back-calculated from the true temperature change and the RC thermal model \cite{hao2014aggregate}, as shown in \eqref{eq:hp_actual_power}. The power command $\hat{P}_{i,t}^{\text{hp}}$ is constrained by \eqref{eq:hp_power_bounds}, and the implied next indoor temperature $\hat{\vartheta}_{i,t+1}$ is constrained by \eqref{eq:hp_temp_bounds}. We also introduce an end-state temperature requirement by \eqref{eq:hp_temp_end}, which ensures thermal flexibility for the next episode. Eqs.~\eqref{eq:hp_cost}--\eqref{eq:hp_net} models the cost for the agent that includes the simplified wear-and-tear cost \eqref{eq:hp_use_cost} and the electricity cost \eqref{eq:hp_energy_cost}.

\subsubsection{Controllable Generator}
The model for a controllable generator is summarised as:
\begin{subequations}\label{eq:gen_model}
    \begin{empheq}[left=\empheqlbrace]{align}
    \Delta\hat{P}_{i,t}^{\text{gen}} &= \hat{P}_{i,t}^{\text{gen}} - P_{i,t-1}^{\text{gen}}, \label{eq:gen_delta_cmd} \\
    \Delta P_{i,t}^{\text{gen}} &= \mathrm{clip} \big(\Delta\hat{P}_{i,t}^{\text{gen}},\,\underline{\Delta P}_i,\,\overline{\Delta P}_i\big), \label{eq:gen_ramp_clip} \\
    P_{i,t}^{\text{gen}} &= P_{i,t-1}^{\text{gen}}+\Delta P_{i,t}^{\text{gen}}, \label{eq:gen_actual_power}
    \end{empheq}
    \begin{empheq}[left=\hspace{-4.75em}\empheqlbrace]{align}
    P_{i,\min}^{\text{gen}} &\le \hat{P}_{i,t}^{\text{gen}} \le P_{i,\max}^{\text{gen}}, \label{eq:gen_bounds} \\
    \underline{\Delta P}_i &\leq \Delta\hat{P}_{i,t}^{\text{gen}} \leq \overline{\Delta P}_i, \label{eq:gen_rampbd}
    \end{empheq}
    \vspace{-2mm}
    \begin{empheq}[left=\empheqlbrace]{align}
    C_{i,t}^{\text{gen}} &= C_{i,t}^{\text{fuel}} + C_{i,t}^{\text{energy}}, \label{eq:gen_cost} \\
    C_{i,t}^{\text{fuel}} &= (a_i P_{i,t}^{\text{gen}} + b_i (P_{i,t}^{\text{gen}})^2)\Delta t, \label{eq:gen_fuel_cost} \\
    C_{i,t}^{\text{energy}} &= \rho_t^{\text{imp}} [p_{i,t}^{\text{net}}]^+ - \rho_t^{\text{exp}} [-p_{i,t}^{\text{net}}]^+, \label{eq:gen_energy_cost} \\
    p_{i,t}^{\text{net}} &= P_{i,t}^{\text{load}} - P_{i,t}^{\text{pv}} - P_{i,t}^{\text{gen}}.
    \label{eq:gen_net}
    \end{empheq}
\end{subequations}
Following the same logic as the battery and heat pump models, $\hat{P}_{i,t}^{\text{gen}}$ denotes the generation setpoint command issued by the agent (control variable), $\Delta\hat{P}_{i,t}^{\text{gen}}$ is the auxiliary \textit{implied power change}, and $P_{i,t}^{\text{gen}}$ denotes the actual generated power after respecting the ramping constraints, achieved through the ramp clipping in \eqref{eq:gen_ramp_clip} based on the lower and upper ramping bounds $\underline{\Delta P}_i$, $\overline{\Delta P}_i$. The actual generated power $P_{i,t}^{\text{gen}}$ is obtained by \eqref{eq:gen_actual_power}. The setpoint command $\hat{P}_{i,t}^{\text{gen}}$ is constrained by \eqref{eq:gen_bounds}, and the implied power change is constrained by \eqref{eq:gen_rampbd}. Eqs.~\eqref{eq:gen_cost}--\eqref{eq:gen_net} model the total agent operation cost as the sum of the quadratic fuel cost and the electricity cost.

Due to the clipping, the per-time-step state constraints are satisfied naturally, so the critical constraints that each agent must learn to satisfy are the end-state requirements in \eqref{eq:bat_soc_end} and \eqref{eq:hp_temp_end}, as well as the network constraints introduced next.

\vspace{-3mm}
\subsection{Three-Phase Network Model}

The household net loads introduced above (device power plus \emph{uncertain} PV and local demand) are connected to different phases and different nodes of a three-phase unbalanced distribution network. Here we assume one household only connects to a single phase. Following standard formulations for unbalanced distribution feeders, we consider a network with one slack bus and $N_B$ three-phase PQ buses, and define:
\begin{itemize}
    \item $\vect{s}_t^Y \in \mathbb{C}^{3N_B}$: Net complex power injections from Wye-connected components at PQ buses.
    \item $\vect{s}_t^D \in \mathbb{C}^{3N_B}$: Net complex power injections from Delta-connected components at PQ buses.
    \item $\vect{v}_t^0 \in \mathbb{C}^3$: Fixed complex voltage at the slack bus.
    \item $\vect{v}_t \in \mathbb{C}^{3N_B}$: Complex nodal voltages at PQ buses.
    \item $\vect{i}_t \in \mathbb{C}^{3N_B}$: Net phase current injections at PQ buses.
    \item $\vect{i}_t^D \in \mathbb{C}^{3N_B}$: Phase-to-phase currents for PQ buses.
\end{itemize}

The electrical characteristics \textcolor{black}{are} represented by the bus admittance matrix $\mat{Y} \in \mathbb{C}^{3(N_B+1)\times 3(N_B+1)}$, partitioned as
\begin{equation}
\mat{Y} := 
\begin{bmatrix} 
\mat{Y}_{00} & \mat{Y}_{0L} \\ 
\mat{Y}_{L0} & \mat{Y}_{LL} 
\end{bmatrix},
\end{equation}
where the sub-matrices denote self and mutual admittances between the slack bus ($0$) and non-slack buses ($L$). 
The AC power flow equations governing steady-state operation are
\begin{subequations}
\begin{align}
    \diag(\mat{H}^\top \conj{\vect{i}_t^D})\vect{v}_t + \vect{s}_t^Y &= \diag(\vect{v}_t)\conj{\vect{i}_t}, \label{eq:pf_wye} \\
    \vect{s}_t^D &= \diag(\mat{H}\vect{v}_t)\conj{\vect{i}_t^D}, \label{eq:pf_delta} \\
    \vect{i}_t &= \mat{Y}_{L0}\vect{v}_t^0 + \mat{Y}_{LL}\vect{v}_t, \label{eq:pf_kirchhoff} \\
     \vect{v}_\text{min} & \leq  |\vect{v}_t| \leq  \vect{v}_\text{max} \label{eq:vol_limit}
\end{align}
\end{subequations}
where $\conj{(\cdot)}$ indicates complex conjugation, and $\mat{H}$ is the transformation matrix relating phase-to-neutral and phase-to-phase quantities (see \cite{3ph_ac_loadflow} for details). Eq.~\eqref{eq:vol_limit} specifies the network voltage magnitude limit, where $|\cdot|$ performs the element-wise norm operation. Real-world distribution networks are typically voltage-constrained and the line current limits can be disregarded \cite{R1D}, especially for distribution networks with long power lines \cite{heinrich2016pv}. Incorporating the line current constraints is also straightforward.

\vspace{-3mm}
\subsection{Multi-Agent Learning Formulation}\label{sec:ma_framework}

Let $\mathcal{N}_{\text{bat}}$, $\mathcal{N}_{\text{hp}}$, and $\mathcal{N}_{\text{gen}}$ denote the battery, heat-pump, and generator agent sets, respectively. Let $\vect{o}_{i,t}$ be the local observation of agent $i$, and let $\hat{P}_{i,t}^{\text{dev}}=\pi_i(\vect{o}_{i,t})$ denote its control action (power command of each of the devices such as $\hat{P}_{i,t}^{\text{gen}}$). We can formulate the multi-agent learning problem through an optimisation lens, which helps us get rid of the Markov restriction in standard RL:
\begin{subequations}\label{eq:policy_opt}
\begin{align}
    \min_{\vect{\theta}:=\{\vect{\theta}_i\}_{i=1}^N} \quad &
    \mathbb{E} \Bigg[
    \sum_{t=1}^{T}
    \Big(
    \sum_{i \in \mathcal{N}_{\text{bat}}} C_{i,t}^{\text{bat}}
    + \sum_{i \in \mathcal{N}_{\text{hp}}} C_{i,t}^{\text{hp}}
    + \sum_{i \in \mathcal{N}_{\text{gen}}} C_{i,t}^{\text{gen}}
    \Big)
    \Bigg] \label{eq:dispatch_obj} \\
    \st \quad & \hat{P}_{i,t}^{\text{dev}} = \pi_i(\vect{o}_{i,t}, \vect{\theta}_i), \quad \forall i,t \label{eq:dispatch_policy} \\
    & \text{battery agents satisfy } \eqref{eq:bat_model}, \quad \forall i \in \mathcal{N}_{\text{bat}} \label{eq:con_bat_model} \\
    & \text{heat-pump agents satisfy } \eqref{eq:hp_model}, \quad \forall i \in \mathcal{N}_{\text{hp}} \label{eq:con_hp_model} \\
    & \text{generator agents satisfy } \eqref{eq:gen_model}, \quad \forall i \in \mathcal{N}_{\text{gen}} \label{eq:con_gen_model} \\
    & \text{with network constraints}\ \text{\eqref{eq:pf_wye}--\eqref{eq:vol_limit}}, \quad \forall t. \label{eq:con_pf}
\end{align}
\end{subequations}
Problem \eqref{eq:policy_opt} is very similar to a standard dispatch problem, except that the decision variables are not the power dispatch actions $\hat{P}_{i,t}^{\text{dev}}$, but the parameterised policies $\{\pi_i(\cdot,\vect{\theta}_i)\}_{i=1}^N$ (such as neural networks (NNs) with parameters $\vect{\theta}_i$) that generate them from local observations \emph{without} online communication. The direct cost summation across all agents in the objective \eqref{eq:dispatch_obj} is appropriate in the present cooperative setting because the goal is to minimise social operating cost, or equivalently maximise social welfare under exogenous prices, and we focus on the fundamental MA dispatch problem rather than fair allocation. The key challenge is not only the network coupling in \eqref{eq:pf_wye}--\eqref{eq:vol_limit}, but also the requirement that good dispatch actions must be generated from local information alone. This is the key difference between our MA framework and standard self-supervised learning approaches \cite{park2024self, chen2025neural, chen2026hard, anrrango2026self}. 

Note that for better learning performance, our agent policies will not directly output device-level power. Instead, let $a_{i,t}$ denote the policy output. It is clipped to $[-1,1]$ and decoded into device-level power by
\begin{align}
    \hat{P}_{i,t}^{\text{bat}} &= a_{i,t} P_{i,\max}^{\text{bat}}, \quad 
    \hat{P}_{i,t}^{\text{hp}} = \frac{a_{i,t}+1}{2} P_{i,\max}^{\text{hp}}, \nonumber \\
    \hat{P}_{i,t}^{\text{gen}} &= P_{i,\min}^{\text{gen}} + \frac{a_{i,t}+1}{2}\big(P_{i,\max}^{\text{gen}}-P_{i,\min}^{\text{gen}}\big),
    \label{eq:action_decode}
\end{align}
which naturally \textcolor{black}{enforces} the device power limits.

\vspace{-1mm}
\section{Gradient-Based Multi-Agent Learning}\label{sec:gradma}

\begin{figure}[t]
    \centering
    \vspace{-1mm}
    \includegraphics[width=1\columnwidth]{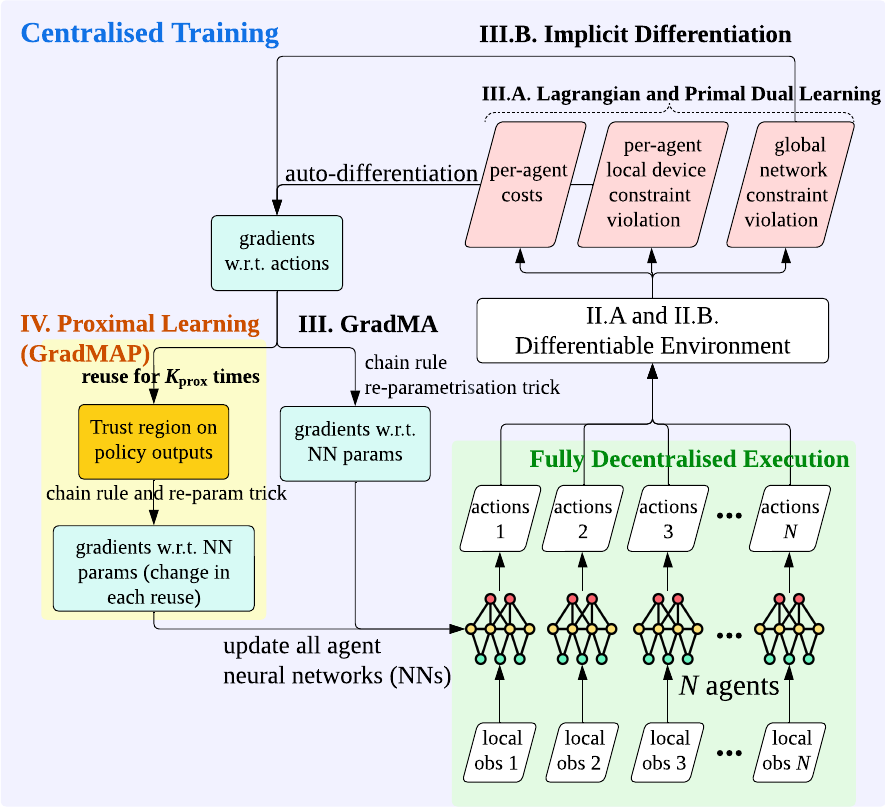}
    \vspace{-6mm}
    \caption{Architecture diagram of the proposed GradMA and GradMAP framework.}
    \label{fig:diagram}
\end{figure}

As discussed in Section \ref{sec:ma_framework}, unlike single-agent approaches \cite{park2024self, chen2025neural, chen2026hard, anrrango2026self}, our parameterised policies $\{\pi_i(\cdot,\vect{\theta}_i)\}_{i=1}^N$ are restricted to local observations for each agent in the online phase. Because of this restriction, we adopt the centralised training and decentralised execution (CTDE) framework \cite{yu2022surprising}, which trains the agent policies with global information but executing them locally, to extend the exact gradient-based self-supervised learning to the MA case (GradMA). Fig.~\ref{fig:diagram} illustrates the architecture of GradMA.

\vspace{-3mm}
\subsection{Exact Gradient-Based Multi-Agent Primal-Dual Learning}\label{sec:exact_grad_pd}

Problem \eqref{eq:policy_opt} is essentially a constrained nonlinear program. A common approach for handling this problem is primal-dual learning \cite{park2024self}, where the hard constraints are first incorporated into the objective to form a Lagrangian loss function. Then gradients (or higher-order information) are used to perform loss minimisation, and Lagrangian multipliers  are gradually increased to impose heavier penalties for constraint violations.

The standard Lagrangian would attach multipliers to every constraint. For the problem scale considered here, this leads to a large dual space and unstable learning dynamics. We therefore aggregate violations into six physically meaningful violation channels in: voltage magnitude $\mathcal{V}_{\text{Volt}}$, battery per-step energy $\mathcal{V}_{\text{Bstp}}$, battery end-state energy $\mathcal{V}_{\text{Bend}}$, heat-pump per-step temperature $\mathcal{V}_{\text{Hstp}}$, heat-pump end-state temperature $\mathcal{V}_{\text{Hend}}$, and generator ramping $\mathcal{V}_{\text{Grmp}}$:
\begin{IEEEeqnarray}{rCl}
    \mathcal{V}_{\text{Volt}}
    & := & \sum_{t=1}^{T}
    \frac{\max(
    [|\vect{v}_t| - \vect{v}_{\max}]^+
    + [\vect{v}_{\min} - |\vect{v}_t|]^+)}{0.5\, (v_{\max} - v_{\min})\, T/N},
    \nonumber\\
    \mathcal{V}_{\text{Bstp}}
    & := & \sum_{t=1}^{T} \sum_{i \in \mathcal{N}_{\text{bat}}}
    \frac{([\hat{E}_{i,t} - E_i^{\max}]^+ + [-\hat{E}_{i,t}]^+)}{P^{\text{bat}}_{i,\max} \, T},
    \nonumber\\
    \mathcal{V}_{\text{Bend}}
    & := & \sum_{i \in \mathcal{N}_{\text{bat}}}
    \frac{[E_{i,\text{target}}^{\text{bat}} - E_{i,T}]^+}{E_{i,\text{target}}^{\text{bat}}},
    \nonumber\\
    \mathcal{V}_{\text{Hstp}}
    & := & \sum_{t=1}^{T} \sum_{i \in \mathcal{N}_{\text{hp}}}\!
    \frac{
    [\hat{\vartheta}_{i,t} - (\vartheta_i^{\text{set}}+\delta_i)]^+
    \! + [(\vartheta_i^{\text{set}}-\delta_i) - \hat{\vartheta}_{i,t}]^+}{\delta_i \, T},
    \nonumber\\
    \mathcal{V}_{\text{Hend}}
    & := & \sum_{i \in \mathcal{N}_{\text{hp}}}
    \frac{[\vartheta_{i,\text{target}}^{\text{hp}} - \vartheta_{i,T}]^+}{\delta_i},
    \nonumber\\
    \mathcal{V}_{\text{Grmp}}
    & := & \sum_{t=1}^{T} \sum_{i \in \mathcal{N}_{\text{gen}}}
    \frac{
    [\Delta\hat{P}_{i,t}^{\text{gen}} - \overline{\Delta P}_i]^+
    + [\underline{\Delta P}_i - \Delta\hat{P}_{i,t}^{\text{gen}}]^+}{0.5 \, (\overline{\Delta P}_i - \underline{\Delta P}_i) \, T}.
    \nonumber
\end{IEEEeqnarray}
Here, $\max(\cdot)$ returns the maximum element of the voltage vector. The denominator in each channel is a normalisation factor that brings the corresponding violation term, when summed over all time steps in a rollout, to a scale comparable to those of the other terms. The network constraint violation term is further scaled \textit{up} by the number of agents so as to preserve its importance in the Lagrangian as the number of agents increases. Although the battery and heat-pump models in Section~\ref{sec:device_model} already enforce per-step state-limit satisfaction through clipping, the violation channels based on the power commands in $\mathcal{V}_{\text{Bstp}}$ and $\mathcal{V}_{\text{Hstp}}$ are empirically found important for effective learning in both gradient-based methods and RL. 

\noindent The rollout Lagrangian is then calculated as follows:
\begin{align}
    &\mathcal{L}(\theta, \vect{\lambda})
     =  \mathbb{E} \Big[
    \Big(\sum\nolimits_{t=1}^{T} C_t^{\text{tot}}\Big)
    + \lambda_{\text{Volt}}\, \mathcal{V}_{\text{Volt}}
    + \lambda_{\text{Bstp}} \mathcal{V}_{\text{Bstp}}+ \nonumber \\
    &\quad\ \lambda_{\text{Bend}} \mathcal{V}_{\text{Bend}}
    + \lambda_\text{Hstp} \mathcal{V}_{\text{Hstp}}
    + \lambda_\text{Hend} \mathcal{V}_{\text{Hend}}
    + \lambda_\text{Grmp}\, \mathcal{V}_{\text{Grmp}}
    \Big]
    \label{eq:lagrangian_loss}
\end{align}
with
\begin{equation}
    C_t^{\text{tot}}
    :=
    \frac{M}{\frac{1}{N}\sum_{i=1}^{N}|P_{i,\max}|}
    (\sum_{i \in \mathcal{N}_{\text{bat}}} C_{i,t}^{\text{bat}}
    + \sum_{i \in \mathcal{N}_{\text{hp}}} C_{i,t}^{\text{hp}}
    + \sum_{i \in \mathcal{N}_{\text{gen}}} C_{i,t}^{\text{gen}}),
    \label{eq:total_stage_cost}
\end{equation}
where the denominator also scales the total cost to make it unitless, so that it is comparable to the constraint-violation terms. 
$M=200$ is a tunable constant that adjusts the importance of the total-cost term in the Lagrangian.

With the Lagrangian defined, the primal-dual learning consists of an outer dual loop with $K_\text{dual}$ steps and an inner primal loop with $K_\text{primal}$ steps. In the inner loop, the policy parameters are updated by gradient descent (or other optimiser such as Adam) on the Lagrangian \eqref{eq:lagrangian_loss}. After the inner primal updates, the dual variables are each updated by projected ascent on the \emph{batch-mean} $(\tilde{\cdot})$ violation channels at a dual learning rate $\alpha$:
\begin{equation}\label{eq:dual_update}
    \lambda_r \leftarrow [\lambda_r + \alpha\, \tilde{\mathcal{V}}_r]^+, \ 
    r \in \{ {\scriptstyle \text{Volt},\, \text{Bstp},\, \text{Bend},\, \text{Hstp},\, \text{Hend},\, \text{Grmp}} \},
\end{equation}
The algorithm stops when the primal-dual gap is sufficiently small or after a maximum number of iterations. Each of the initial dual values $\lambda_r$ can be set to zero.

\subsection{Implicit Differentiation for AC Power Flow}
\vspace{-1mm}

As illustrated in Fig.~\ref{fig:diagram}, a core part in the primal-dual training process is the evaluation of gradients of the Lagrangian  \eqref{eq:lagrangian_loss}. Most of the gradient terms can be obtained through automatic differentiation (AD) available in modern deep learning frameworks. The AC network constraint part is different: the voltage profile is obtained from an iterative three-phase power-flow solve, so naively applying AD would unroll all solver iterations and backpropagate through the entire loop, which creates computational and memory issues. Instead, we use implicit differentiation on the converged fixed point.

Using the Z-Bus formulation \cite{zbus}, let $\mat{Z} = \mat{Y}_{LL}^{-1}$ and $\vect{w} = -\mat{Z}\mat{Y}_{L0}\vect{v}_t^0$. Then the power flow can be written as
\begin{equation}
    \vect{v}_t
    =
    \Phi(\vect{v}_t,\vect{s}_t)
    :=
    \mat{Z}\vect{i}_{\text{inj}}(\vect{v}_t,\vect{s}_t) + \vect{w},
    \label{eq:fixed_point}
\end{equation}
where $\vect{s}_t$ stacks the Wye and Delta load injections and $\vect{i}_{\text{inj}}(\cdot)$ is the corresponding current injection map implied by \eqref{eq:pf_wye}--\eqref{eq:pf_delta}. At the converged solution $\vect{v}_t^\star=\Phi(\vect{v}_t^\star,\vect{s}_t)$, differentiating both sides w.r.t. the load injections gives
\begin{equation}
    \Big(\mat{I}-\frac{\partial \Phi}{\partial \vect{v}_t}(\vect{v}_t^\star,\vect{s}_t)\Big)
    \frac{\partial \vect{v}_t^\star}{\partial \vect{s}_t}
    =
    \frac{\partial \Phi}{\partial \vect{s}_t}(\vect{v}_t^\star,\vect{s}_t),
    \label{eq:implicit_grad}
\end{equation}
The term $\frac{\partial \vect{v}_t^\star}{\partial \vect{s}_t}$ is the Jacobian matrix of the converged voltage w.r.t. the load injections. Explicitly computing this full Jacobian matrix is computationally expensive and unnecessary. To update the policies, the actual quantity we target to evaluate is the gradient of the voltage-violation \textit{scalar} term w.r.t. the load injection \textit{vector} at time step $t$, namely $\frac{\partial \mathcal{V}_{\text{Volt}}}{\partial \vect{s}_t}$. Let
\(
\vect{g}_v := \frac{\partial \mathcal{V}_{\text{Volt}}}{\partial \vect{v}_t^\star}
\),
which can be obtained directly by AD. Then, by the chain rule and \eqref{eq:implicit_grad}, we can re-write our target gradient as
\begin{align}
    \frac{\partial \mathcal{V}_{\text{Volt}}}{\partial \vect{s}_t}
    &=
    \Big(\frac{\partial \vect{v}_t^\star}{\partial \vect{s}_t}\Big)^\top \vect{g}_v \nonumber\\
    &=
    \Big(\frac{\partial \Phi}{\partial \vect{s}_t}(\vect{v}_t^\star,\vect{s}_t)\Big)^\top
    \Big(\mat{I}-\frac{\partial \Phi}{\partial \vect{v}_t}(\vect{v}_t^\star,\vect{s}_t)\Big)^{-\top}
    \vect{g}_v.
    \label{eq:vector_jacobian}
\end{align}
If we define a \emph{vector}
\(
\vect{\gamma}_t
:=
\big(\mat{I}-\frac{\partial \Phi}{\partial \vect{v}_t}(\vect{v}_t^\star,\vect{s}_t)\big)^{-\top}
\vect{g}_v
\),
then we can calculate $\vect{\gamma}_t$ by solving
\begin{equation}
    \Big(\mat{I}-\big(\frac{\partial \Phi}{\partial \vect{v}_t}(\vect{v}_t^\star,\vect{s}_t)\big)^\top\Big)\vect{\gamma}_t = \vect{g}_v,
    \label{eq:adjoint_pf}
\end{equation}
Then our target gradient is calculated as
\begin{equation}
    \frac{\partial \mathcal{V}_{\text{Volt}}}{\partial \vect{s}_t}
    = \Big(\frac{\partial \Phi}{\partial \vect{s}_t}(\vect{v}_t^\star,\vect{s}_t)\Big)^\top \vect{\gamma}_t.
    \label{eq:sensitivity_s}
\end{equation}

\vspace{-3mm}
\subsection{Re-parameterisation Trick for Stochastic Policy Learning}

Standard self-supervised gradient methods are usually presented with deterministic policies \cite{park2024self, chen2025neural, chen2026hard, anrrango2026self}. In practice, however, deterministic policies can make the optimisation landscape brittle and can get trapped in poor local minima. To handle this issue, we introduce stochastic policies that enable exploration and smoothing, and optimise them through the re-parameterisation trick \cite{son2023gradient}. More specifically, for each agent, the policy outputs the parameters of a Gaussian distribution over a normalised action,
\begin{equation}
    \pi_{\vect{\theta}_i}(\cdot \mid \vect{o}_{i,t})
    =
    \mathcal{N}\!\big(
    \mu_{\vect{\theta}_i}(\vect{o}_{i,t}),
    \sigma_{\vect{\theta}_i}^2(\vect{o}_{i,t})
    \big),
    \label{eq:gaussian_policy}
\end{equation}
Given $\vect{o}_{i,t}$, the sampled action is
\begin{equation}
    a_{i,t}
    =
     \mu_{\vect{\theta}_i}(\vect{o}_{i,t})
    +
    \sigma_{\vect{\theta}_i}(\vect{o}_{i,t})\, \epsilon_{i,t},
    \quad
    \epsilon_{i,t} \sim \mathcal{N}(0,1).
    \label{eq:reparam}
\end{equation}
$a_{i,t}$ is then clipped to $[-1,1]$ and decoded into device-level power according to \eqref{eq:action_decode}. If $\partial \mathcal{L} / \partial a_{i,t}$ denotes the action gradient returned by the differentiable environment, then the gradients in policy-output coordinates are
\begin{equation}
    \frac{\partial \mathcal{L}}{\partial \mu_{i,t}}
    =
    \frac{\partial \mathcal{L}}{\partial a_{i,t}},
    \qquad
    \frac{\partial \mathcal{L}}{\partial \sigma_{i,t}}
    =
    \frac{\partial \mathcal{L}}{\partial a_{i,t}} \epsilon_{i,t},
    \label{eq:mu_sigma_grad}
\end{equation}
which can then be used for policy optimisation.

\vspace{-2mm}
\section{Gradient-Based Proximal Learning}\label{sec:gradmap}

In exact-gradient-based GradMA (and also existing self-supervised learning approaches), after each differentiable environment rollout, fresh gradient is computed and updates the policy only once. However, differentiating the environment particularly with the implicit differentiation for the three-phase AC power flow can be dominant computational bottlenecks. This section thus introduces GradMAP, which first collects rollouts, computes the corresponding Lagrangian gradients w.r.t. agents' policy outputs (rather than policy parameters), and then \textit{reuses} these cached gradients for multiple inner updates within a trust region in the policy-output space, as illustrated in Fig.~\ref{fig:diagram}.

Suppose for each primal step a rollout batch has been generated by an old policy with outputs $(\mu_{i,t}^{\text{old}}, \sigma_{i,t}^{\text{old}})$, and let $(g^{\mu}_{i,t}, g^{\sigma}_{i,t})$ be the corresponding environment gradients w.r.t. the policy outputs $(\mu_{i,t}^{\text{old}}, \sigma_{i,t}^{\text{old}})$ via \eqref{eq:mu_sigma_grad}. GradMAP tries to optimise the following problem in each primal step:
\begin{align}\label{eq:constrained_gradmap}
    \min_{\vect{\theta}:=\{\vect{\theta}_i\}}\ \mathbb{E} \Big[
    &g^{\mu}_{i,t} \mu_{i,t}(\bm{\theta}_i)
    + g^{\sigma}_{i,t} \sigma_{i,t}(\bm{\theta}_i) 
    - \tau \mathcal{H}(\pi_{\bm{\theta}_i}(\cdot \mid \vect{o}_{i,t}))
    \Big]\nonumber \\
    &\text{s.t.}\ \mathcal{T}(\vect{\theta})\le \epsilon_{\mathrm{tr}},
\end{align}
where $\tau \mathcal{H}(\pi_{\theta}(\cdot \mid \vect{o}_{i,t}))$ is an entropy bonus to encourage exploration, which is commonly used in stochastic policy learning \cite{yu2022surprising}. It is apparent that, without the trust region and the entropy bonus, differentiating \eqref{eq:constrained_gradmap} w.r.t. policy outputs (not $\vect{\theta}$) gives exactly the gradients to be reused $(g^{\mu}_{i,t}, g^{\sigma}_{i,t})$. The trust region is defined as:
\begin{equation}
    \mathcal{T}(\vect{\theta}) :=
    \big(
    \mathbb{E} \big[
    \|\mu_{i,t}(\bm{\theta}_i)-\mu_{i,t}^{\text{old}}\|_2^2 +
    \|\sigma_{i,t}(\bm{\theta}_i)-\sigma_{i,t}^{\text{old}}\|_2^2
    \big]
    \big)^{1/2},
    \label{eq:trust_metric}
\end{equation}
where $\mathbb{E}$ represents the average over all agents, all time steps, and all rollouts in the batch (this is feasible because we have assumed centralised offline training as part of CTDE).
{\color{black}
The trust region is defined directly in the policy-output space $(\mu,\sigma)$ (we use squared distance, although other distance metrics in policy-output space are also possible), rather than indirectly limiting policy updates through probability ratios or KL divergence in distribution space, as in standard PPO. This output-space formulation is a more natural choice here because the reused environment gradients are defined exactly with respect to the policy outputs. More formally, the GradMAP surrogate uses a first-order approximation in these same output coordinates, so the proximal term can more directly penalise movement away from the expansion point and helps keep updates local.
In implementation, we do not solve the constrained problem directly. Instead, the trust-region constraint is converted into a \textcolor{black}{quadratic regularisation term}, yielding the GradMAP surrogate loss:}
\begin{align}
    \ell_{\text{GMAP}}\!
    =\! \mathbb{E}& \Big[
    g^{\mu}_{i,t} \mu_{i,t}(\bm{\theta}_i)
    + g^{\sigma}_{i,t} \sigma_{i,t}(\bm{\theta}_i) - \tau \mathcal{H}(\pi_{\bm{\theta}_i}(\cdot \mid \vect{o}_{i,t})) \nonumber \\
    &+\! \frac{\beta}{2}(\|\mu_{i,t}(\bm{\theta}_i)\!-\!\mu_{i,t}^{\text{old}}\|_2^2
    \!+\! \|\sigma_{i,t}(\bm{\theta}_i)\!-\! \sigma_{i,t}^{\text{old}}\|_2^2) \Big],
    \label{eq:gmap_loss}
\end{align}
where $\beta$ is an adaptively scaled trust-region penalty coefficient. We can then perform $K_\text{prox}$ updates to minimise $\ell_{\text{GMAP}}$. Note that in the $K_\text{prox}$ updates, although the same environment gradients w.r.t. the agent policy outputs are re-used, the gradients w.r.t. the agent NN parameters are changing due to both the changing trust region penalty and the update in NN parameters.

After each primal step we average $\mathcal{T}(\vect{\theta})$ and then update $\beta$ by dynamic multiplicative scaling (following the KL-penalty version of PPO \cite{ppo2017proximal}):
\begin{equation}\label{eq:beta_adapt}
    \beta \leftarrow
    \begin{cases}
        1.1\,\beta, & \mathcal{T}(\vect{\theta}) > \epsilon_{\mathrm{tr}},\\
        \beta/1.1, & \mathcal{T}(\vect{\theta}) < \epsilon_{\mathrm{tr}}/2,\\
        \beta, & \text{otherwise},
    \end{cases}
    \qquad
    \beta \in [50,10^4].
\end{equation}
Algorithm~\ref{alg:gradmap} summarises the triple-loop structure of GradMAP: an outer dual loop for constraint penalty updates, a middle primal loop under fixed dual variables, and a fast innermost surrogate-optimisation loop. The innermost updates are computationally cheap because of the gradient reuse.

{\color{black} 
\begin{remark}
Note that KL remains a valid metric because it still implicitly restricts the policy-output displacement, although it leads to worse performance than our more direct policy-output regularisation, as demonstrated in \nameref{app:ablation}.
\end{remark}
}

{\color{black} 
\begin{remark}
GradMAP improves the computational efficiency of multi-agent learning but does not, by itself, guarantee constraint satisfaction. It can be combined with a range of safety mechanisms, as can standard MAPPO \cite{su2025review}. The primal-dual formulation adopted here provides a soft mechanism for constraint handling, while hard safety mechanisms, such as safety layers and shielding, can also be incorporated.
\end{remark}

\begin{remark}
GradMAP's policy-output gradient-reuse mechanism is agnostic to network structure or information-sharing or parameter-sharing mechanisms, such as graph neural networks, mean-field approximations, or parameter-sharing actors. The key feature of GradMAP is the reuse of exact environment gradients within a policy-output-space trust region, whereas standard MARL methods such as MAPPO estimate policy gradients from sampled trajectories.
\end{remark}
}

\begin{algorithm}[!t]
\caption{GradMAP training procedure}
\label{alg:gradmap}
\begin{algorithmic}[1]
\STATE Initialise all agents' policy parameters $\{\vect{\theta}_i\}$, dual variables $\vect{\lambda}$, and trust coefficient $\beta$
\FOR{$m = 1,\dots,K_{\text{dual}}$}
\FOR{$j = 1,\dots,K_{\text{primal}}$}
\STATE Interact with the environment and collect a batch of rollouts
\STATE Store rollout statistics $(\mu^{\text{old}}, \sigma^{\text{old}})$ and differentiate the rollout Lagrangian \eqref{eq:lagrangian_loss} w.r.t. actions
\STATE Convert action gradients into cached $(g^{\mu}, g^{\sigma})$ using the re-parameterisation rule \eqref{eq:mu_sigma_grad}
\FOR{$e = 1,\dots,K_{\text{prox}}$}
\STATE Update ${\vect{\theta}_i}$ of all agents by minimising \eqref{eq:gmap_loss} using a deep learning optimiser (e.g. Adam)
\ENDFOR
\STATE Update $\beta$ according to \eqref{eq:beta_adapt}
\ENDFOR
\STATE Update each dual multiplier by \eqref{eq:dual_update}
\ENDFOR
\end{algorithmic}
\end{algorithm}

\begin{figure}[tb]
    \centering
    \vspace{-4mm}
    \includegraphics[width=1\columnwidth]{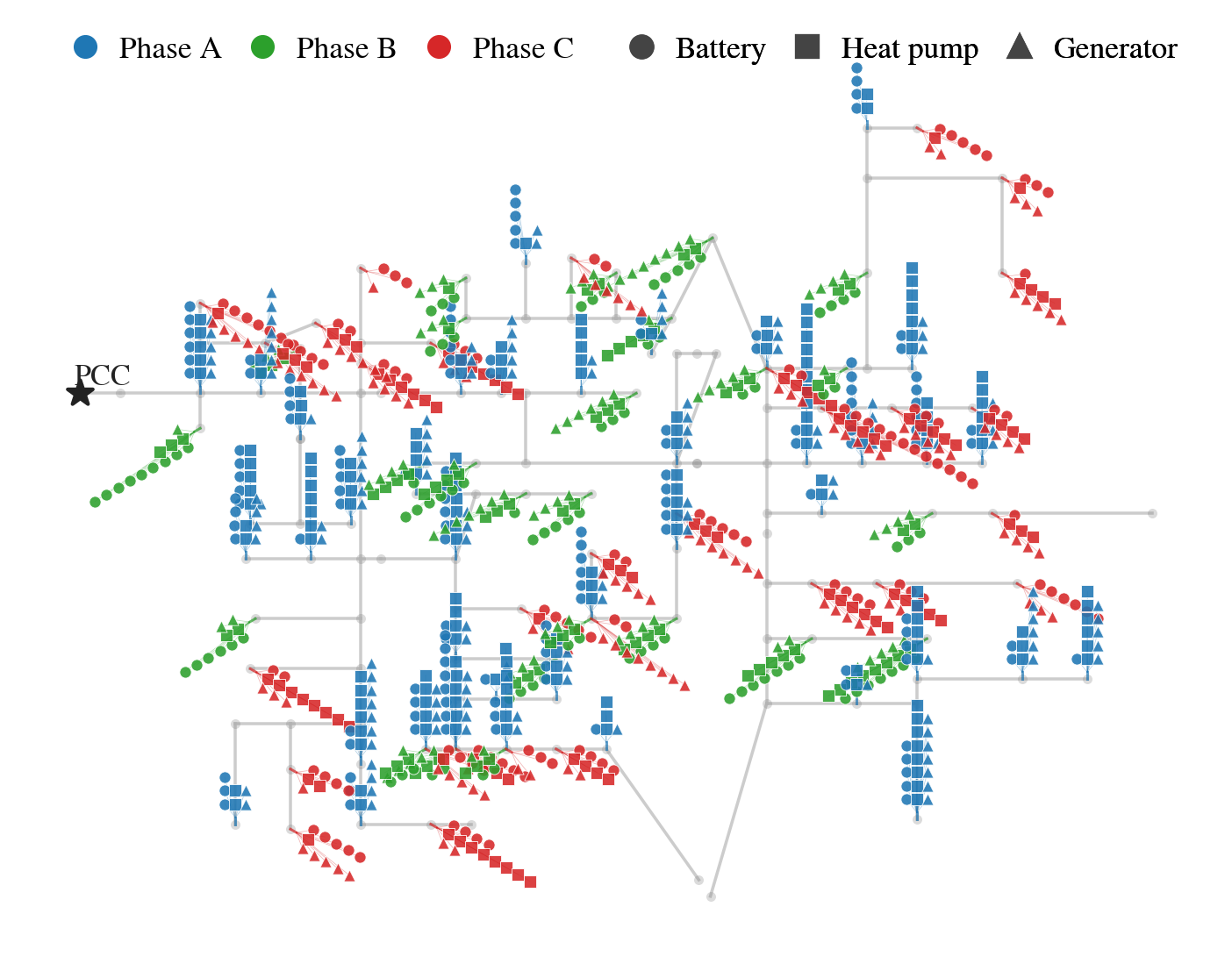}
    \vspace{-12mm}
    \caption{IEEE123 feeder topology used in the main case study together with the connection locations of the $1{,}000$ agents. Gray nodes denote feeder buses, coloured branches indicate occupied phase connections, and marker types distinguish the $334$ battery, $333$ heat-pump, and $333$ generator agents distributed across the network.}
    \label{fig:feeders}
\end{figure}

{\color{black}
\begin{figure}[tb]
    \centering
    \vspace{-4mm}
    \includegraphics[width=1\columnwidth]{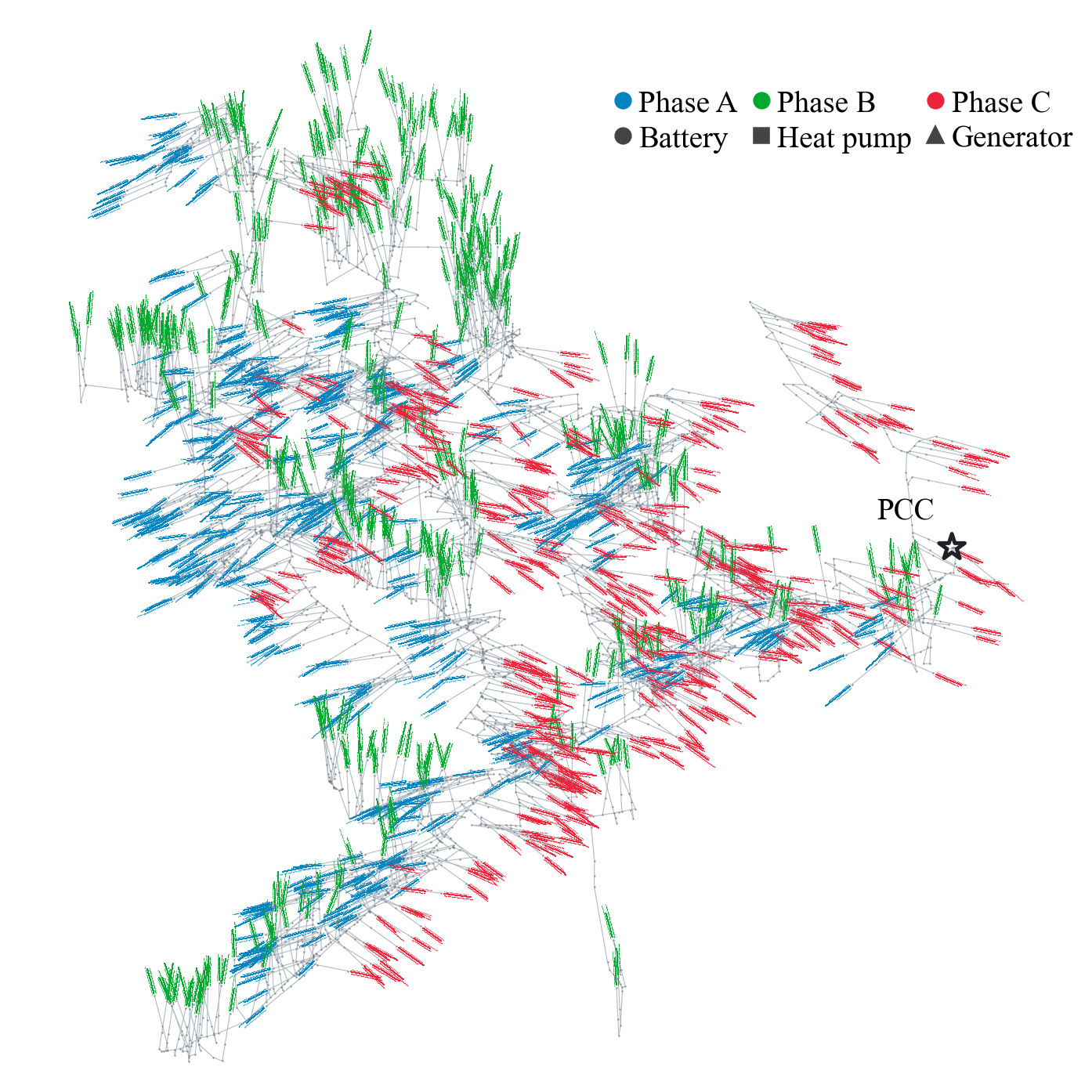}
    \vspace{-8mm}
    \caption{\color{black}IEEE8500 feeder topology used in the main case study together with the connection locations of the $100{,}000$ agents. Gray nodes denote the $4{,}876$ feeder buses (since low-voltage bus coordinates are not provided in the feeder data, these buses are positioned schematically around their upstream medium-voltage buses for clarity). Coloured branches indicate occupied phase connections, and marker types distinguish the $33{,}334$ battery, $33{,}333$ heat-pump, and $33{,}333$ generator agents distributed across the network.}
    \label{fig:feeders8500}
\end{figure}
}

\begin{figure}[t]
    \centering
    \vspace{-0mm}
    \includegraphics[width=1\columnwidth]{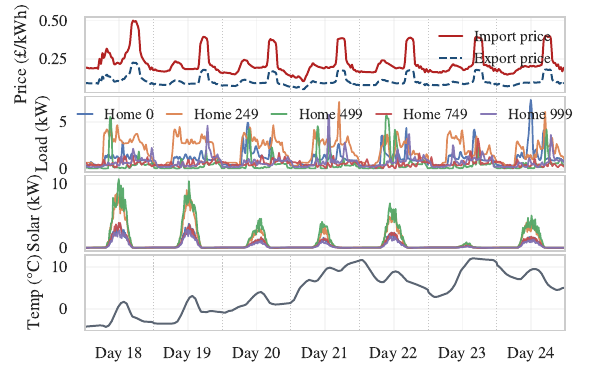}
    \vspace{-10mm}
    \caption{Seven-day visualisation of the datasets used in the IEEE123 $1{,}000$-agent case study.}
    \label{fig:data}
\end{figure}

\section{Case Studies}\label{sec:case_study}


\begin{figure}[tb]
    \centering
    \includegraphics[width=\columnwidth]{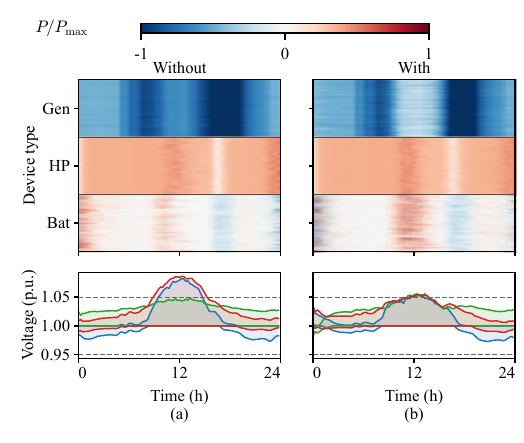}
    \vspace{-10mm}
    \caption{{\color{black}All 1,000 agents' normalised power on held-out test day 18: (a) without and (b) with network constraint signals. Each row corresponds to the power profile of one agent. Positive values represent demand and negative values generation. Voltage envelopes cover all valid buses: phase A (blue), B (green) and C (red). Grey dashed lines mark the preset voltage limit $0.95$--$1.05$ p.u. The phase voltage of this day ranges are $0.9728$--$1.0827$, $0.9998$--$1.0488$ and $0.9894$--$1.0865$ p.u. in (a), and $0.9725$--$1.0560$, $0.9820$--$1.0561$ and $0.9887$--$1.0549$ p.u. in (b).}}
    \label{fig:all_agent_dispatch_1000}
\end{figure}

\begin{figure}[tb]
    \centering
    \includegraphics[width=\columnwidth]{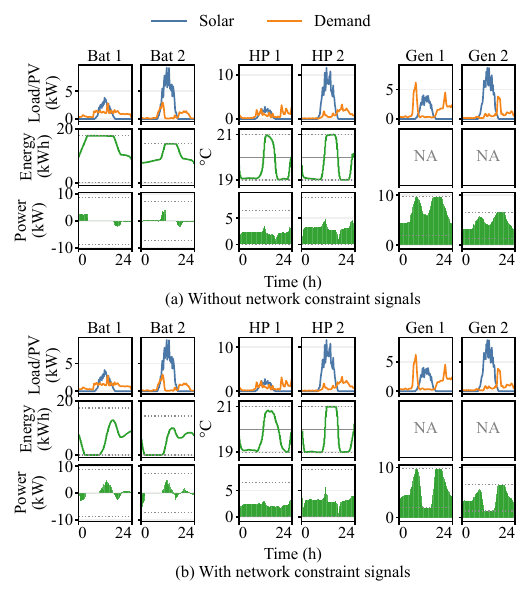}
    \vspace{-8mm}
    \caption{{\color{black}Zoomed-in illustration for Fig.~\ref{fig:all_agent_dispatch_1000}, showing the same two agents per device type: (a) without and (b) with network constraint signals. Bat, HP and Gen denote batteries, heat pumps and generators. Rows show load/PV, device state and power. Grey dotted lines mark power/energy limits, and solid lines mark temperature set points.}}
    \label{fig:single_day_rollout_1000}
\end{figure}

\vspace{-1mm}
\subsection{Experimental Setup}\label{sec:exp_setup}
\vspace{-1mm}

To demonstrate the performance of GradMAP, we present a case study based on the IEEE 123-bus mid-voltage feeder \cite{ieee_feeder} with 1{,}000 agents. For scaling comparisons we additionally present case studies with the IEEE 13-bus medium-voltage feeder with 10 agents, the reduced EULV 124-bus low-voltage feeder with 100 agents, and IEEE8500 distribution feeder with 100,000 agents \cite{ieee_feeder}. Fig.~\ref{fig:feeders} presents the IEEE 123-bus network together with the connection points of the 1{,}000 agents used in the main study. For training and testing, three-phase unbalanced AC power flow is solved via the Z-Bus method \cite{zbus}. 

{\color{black} To evaluate the voltage gradient, we solve the linear system \eqref{eq:adjoint_pf} using the Krylov iterative method BiCGSTAB, because it is natively available in JAX and avoids explicitly forming the coefficient matrix, thus reducing memory usage. A trade-off is the possibility of convergence issues that lead to a small number of NaN network gradients at the start of training. We set these few NaNs to zero without impacting training. Preconditioning or a direct linear-system solve (such as an LU- or QR-based solve) can be used to improve robustness.}

Real-world time-series datasets are used in this paper. The non-controllable household load data are obtained from the residential measurement dataset in \cite{wardle2020dataset}, from which we extract 2{,}000 half-hourly household demand time series spanning 15 months after preprocessing. The hourly solar-generation and air-temperature time series are sourced from \cite{renewable_ninja}. The import and export price data are obtained from the UK Octopus Energy Agile tariff \cite{octopus_agile_historical_data}, a residential tariff scheme that varies on a half-hourly and daily basis. Fig.~\ref{fig:data} illustrates these data over seven consecutive days. Linear interpolation is used to increase the temporal resolution. In our case studies, each time step is 15 min, and each episode spans one day with a 24-h horizon, giving $T=96$. 

All decentralised policies use the same 8-dimensional local observation at the \emph{current} time step, defined as
\begin{equation}
    \vect{o}_{i,t}
    =
    [
    \bar{t},\,
    \bar{x}_{i,t},\,
    \bar{d}_{i,t}^{\mathrm{net}},\,
    \bar{p}_t^{\mathrm{imp}},\,
    \bar{p}_t^{\mathrm{exp}},\,
    \bar{\theta}_t^{\mathrm{out}},\,
    \bar{u}_{i,t},\,
    \bar{v}_{i,t}^{\mathrm{loc}}
    ]^{\top},
    \label{eq:local_obs}
\end{equation}
where the entries are the normalised time $\bar{t}:=t/T$, device state $\bar{x}_{i,t}$ (energy level for batteries, temperature for heat pumps, and power at the previous time step for generators), net uncontrollable demand $\bar{d}_{i,t}^{\mathrm{net}}$ (uncontrollable load minus solar generation), import and export prices $\bar{\rho}_t^{\mathrm{imp}}$ and $\bar{\rho}_t^{\mathrm{exp}}$, outdoor temperature $\bar{\theta}_t^{\mathrm{out}}$, end-state urgency $\bar{u}_{i,t}$, and local voltage $\bar{v}_{i,t}^{\mathrm{loc}}$. All features are normalised to the interval $[0,1]$ using either the device parameters or the training data statistics. The end-state urgency feature $\bar{u}_{i,t}$ was found to be helpful for satisfying the end-state constraints. It is calculated as:
\begin{equation}
    \label{eq:urgency_term}
    \begin{aligned}
        \bar{u}_{i,t}
        =
        \frac{\mathbf{1}_{\{i \in \mathcal{N}_{\text{bat}}\}}(E_{i,\text{target}}^{\text{bat}} - E_{i,t})}{(1-\bar{t}) (E_i^{\max}/2)} +
        \frac{\mathbf{1}_{\{i \in \mathcal{N}_{\text{hp}}\}}(\vartheta_{i,\text{target}}^{\text{hp}} - \vartheta_{i,t})}{(1-\bar{t}) \delta_i}.
    \end{aligned}
\end{equation}
{\color{black} 
These eight features should be available to local agents through standard sensors and their energy suppliers' public APIs. The end-state urgency is an artificial feature that only requires the user's pre-specified end-state requirements.
}

The reported performance metrics are the total operating cost \eqref{eq:dispatch_obj}, the voltage-violation magnitude, the battery and heat-pump end-state violation magnitudes, the training wall-clock time, and the number of primal steps to reach specific performance. 
We evaluate the proposed GradMAP against several benchmarks. To highlight the advantage of our proposed gradient reuse with a policy-output-space trust region, we include GradMA-RPO that adapts the single-agent PPO-style probability-distribution-space trust region from \cite{zhong2025reparameterization}. We also compare against GradMA (Section~\ref{sec:gradma}), our proposed MA extension of existing self-supervised learning methods \cite{park2024self, chen2025neural, chen2026hard, anrrango2026self}. Finally, we include standard black-box MARL algorithms, namely IPPO and MAPPO \cite{yu2022surprising}.
Following the MAPPO implementation from \cite{yu2022surprising}, all agents share a single global critic network. During offline training, this shared critic receives the full set of observations from all agents, and returns an agent-specific value estimate. Note that \cite{yu2022surprising} also proposed using the global state together with each agent's local observation rather than concatenating all observations, in order to maintain computational scalability. However, our empirical tests show that this performs worse. For this reason, our MAPPO can only be tested in the 10-agent and 100-agent cases, but is not computationally feasible in the 1{,}000-agent case. Another computationally efficient variant of MAPPO outputs the same value estimate for all agents \cite{yu2022surprising}, but we found this performs even worse than IPPO.

We also include a naive baseline to demonstrate that we achieve meaningful learning outcomes. In this baseline, the batteries remain idle, the heat pumps always maintain the indoor temperature at the reference level, and the generators always operate at 100\% power.

\textcolor{black}{We use 12 consecutive months of data for training and hyperparameter tuning. The last month is used as a held-out validation set to determine the hyperparameters, after which all methods are re-trained over the full 12 months before testing.
The subsequent month, January (31 days), is used for out-of-sample testing. During testing, the days are treated as consecutive, such that the final state of one day is used as the initial state of the next day. Table~\ref{tab:hyperparams} summarises the resulting hyperparameter settings. All methods use primal--dual learning, batches of 500 episodes per primal step, and the Adam optimiser. GradMA uses a larger learning rate because it evaluates the exact gradient at each update. The policy network for each agent is a small NN with a single hidden layer of 16 neurons (so two layers in total) and tanh activation, which performs better empirically than deeper or wider fully-connected networks. Although each network is small, it should be noted that there are many (up to 100{,}000) independent agents in total with no parameter sharing. The number of training iterations for each method is chosen based on the convergence of the cost and constraint violations on the validation set. Compared with GradMAP and GradMA-RPO, GradMA requires more primal steps to converge because it does not reuse gradient information. IPPO and MAPPO reuse environment rollout data multiple times with clipping, but their sample efficiency is lower; we therefore increase both their number of dual steps ($K_\text{dual}$) and primal steps ($K_\text{primal}$). In addition, because IPPO and MAPPO are sample-based methods, their initial policy standard deviation is set to $1$ (0 in log scale), which is a standard choice and is observed to have better validation performance. The other methods have gradient information available and can therefore use a smaller initial standard deviation of approximately 0.14 (-2 in log scale), which also leads to better validation performance.}

To improve model performance for consecutive testing, during training we randomly scale the initial states of all grid-edge devices. The script is written in JAX 0.4.38 on Python 3.10.19, and all experiments are run on a single NVIDIA RTX PRO 5000 Blackwell 48 GB GPU (non-data-centre GPU).

\begin{table}[tb]
\vspace{-2mm}
    \caption{Main training hyperparameters.}
    \vspace{-3mm}
    \label{tab:hyperparams}
    \centering
    \scriptsize
    \setlength{\tabcolsep}{3pt}
    \renewcommand{\arraystretch}{1.05}
    \resizebox{\columnwidth}{!}{%
    \begin{tabular}{lccccc}
        \hline
        Hyperparameter & GradMAP & GradMA-RPO & GradMA & IPPO & MAPPO \\
        \hline
        Actor NN & 2 layers with 16 neurons & same & same & same & same \\
        Critic NN & N/A & N/A & N/A & 2$\times$64 & 2$\times$64 \\
        $K_{\text{dual}}$ & 20 & 20 & 20 & 40 & 40 \\
        $K_{\text{primal}}$ & 10 & 10 & 60 & 30 & 30 \\
        $K_{\text{prox}}$ & 80 & 80 & N/A & 80 & 80 \\
        NN learning rate & $5{\times}10^{-4}$ & $5{\times}10^{-4}$ & $2{\times}10^{-3}$ & $5{\times}10^{-4}$ & $5{\times}10^{-4}$ \\
        Dual lr $\alpha$ & 150 & 150 & 150 & 150 & 150 \\
        Init.\ std & 0.14 & 0.14 & 0.14 & 1.0 & 1.0 \\
        Entropy coeff. & 0.01 & 0.01 & 0.01 & 0.005 & 0.005 \\
        \shortstack[c]{Trust / clipping\\ \ } & \shortstack[c]{adaptive prox\\$\mathcal{T}(\vect{\theta})\le 0.03$} & \shortstack[c]{ratio gate $[0.2,2.0]$\\+ KL 0.2} & none & PPO clip 0.2 & PPO clip 0.2 \\
        \hline
    \end{tabular}%
    }
\end{table}

\vspace{-3mm}
\subsection{Single-Day Dispatch}

To illustrate the effectiveness of the proposed GradMAP in handling AC network constraints, Fig.~\ref{fig:all_agent_dispatch_1000} shows the normalised power trajectories for all the 1,000 agents in a single representative day in both the unconstrained case (where the voltage-violation signal is set to zero during training) and the constrained case. As shown, without the voltage-violation signal, the agents learn solely from the price signal, and reduce demand or increase discharge/generation during the afternoon price peak (see Fig. \ref{fig:data} Day 18). In particular, the generators produce power profiles that closely follow the price signal because of the quadratic cost structure.
In contrast, when the constraint signal is included during training, GradMAP effectively reduces the maximum phase-voltage violation from 0.0365~p.u. to 0.0061~p.u. by reducing generator output and/or increasing battery charging during constrained periods (a clear upward shift in the right panel of Fig.~\ref{fig:all_agent_dispatch_1000}).

Fig.~\ref{fig:single_day_rollout_1000} also provides a zoomed-in visualisation of selected agents. The overall trend is similar. It is noteworthy that one heat pump (HP1) instead reduces its power consumption, possibly due to the three-phase network coupling effect. Satisfying the network constraints also involves a cost trade-off, with the total cost increasing from £5{,}027 to £5{,}219 on this day.

\begin{figure}[tb]
    \centering
    \includegraphics[width=\columnwidth]{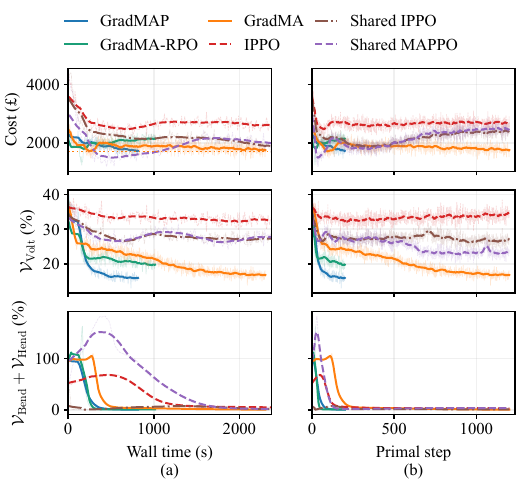}
    \vspace{-8mm}
    \caption{{\color{black}Training convergence plots for the $1{,}000$-agent case. The primal step corresponds to $K_\text{primal}$ in Algorithm~\ref{alg:gradmap}. Each primal step represents one batched environment-rollout interaction. The horizontal dotted line marks GradMA's lowest smoothed training cost. Darker curves show 30-step rolling averages and lighter curves show batch values.}}
    \label{fig:training_convergence_1000}
\end{figure}

\vspace{-3mm}
\subsection{Training Convergence}
\vspace{-1mm}

Fig.~\ref{fig:training_convergence_1000} compares the proposed GradMAP with other learning-based benchmarks in terms of training convergence on the same 1{,}000-agent case. GradMAP not only converges faster, within \textit{only 15 minutes}, but also converges to a better solution than all other benchmarks in terms of both operating cost and constraint satisfaction. IPPO is  unable to learn effectively in this setting, likely because its critic networks only take local observation and cannot capture the complex network coupling.
Although GradMA-RPO also shows fast convergence in constraint satisfaction, its operating cost starts to increase after around 100 primal steps, indicating an inability to effectively balance constraint satisfaction and cost minimization. This supports our argument that our proposed policy-output-space trust region in GradMAP is more principled and therefore leads to better learning performance than the PPO-style trust region defined in the probability distribution space \cite{zhong2025reparameterization} \textcolor{black}{Fig.~\ref{fig:penalty-comparison} in \nameref{app:ablation} provides further evidence}. The exact-gradient GradMA is the only method that achieves operating-cost and constraint performance comparable to that of GradMAP, but it is much slower in both wall-clock time and primal steps. In particular, focusing on the operating cost in the first row of plots, GradMAP is approximately 3--5$\times$ faster in the wall-clock time than GradMA in converging to the same cost level.

\vspace{-3mm}
\subsection{Out-of-Sample Benchmark Comparison}
\vspace{-0mm}

\begin{figure}[tb]
    \centering
    \vspace{-2mm}
    \includegraphics[width=\columnwidth]{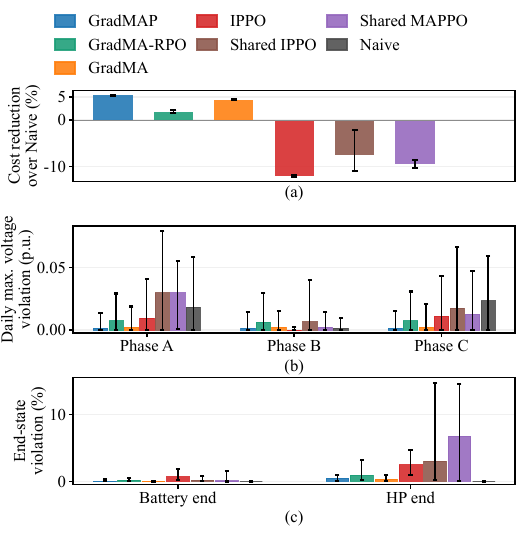}
    \vspace{-8mm}
    \caption{{\color{black}Out-of-sample benchmark comparison for the $1{,}000$-agent case over 31 consecutive test days. (a) Cost reduction relative to the Naive method from 31-day totals (higher value indicates lower cost): average (the main bar) and min--max (the error bar) across three different training seeds. (b,c) Averages (the main bar) and min--max (the error bar) ranges of 93 daily values (31 for Naive). Voltage uses the daily maximum over buses and time steps within each phase.}}
    \label{fig:oos_comparison_1000}
\end{figure}

Fig.~\ref{fig:oos_comparison_1000} compares the out-of-sample performance over 31 consecutive held-out days, where all learning methods are trained and evaluated under three different random seeds to improve the robustness of the conclusions. {\color{black}
To assess performance against more scalable MARL baselines, we also include two parameter-sharing variants: Shared IPPO and Shared MAPPO. In each variant, agents of the same device type share an actor network and a critic network, with two additional input features representing the device power capacity and household PV capacity. Across all the methods compared and all three seeds, GradMAP achieves the greatest cost reduction relative to the naive baseline while also keeping the mean and worst-case values of daily maximum voltage violations and daily end-state violations among the lowest. Its advantage over the MARL baselines is evident in both the means and the upper endpoints of the min--max ranges.
}

\vspace{-3mm}
\subsection{Scalability Study}
\vspace{-0mm}

{\color{black}
Fig.~\ref{fig:scale_trends} compares GradMAP and other benchmarks across different problem scales up to 100,000 agents (with no parameter sharing, giving 16.2 million parameters in total) on the IEEE8500 network \cite{ieee_feeder}. Due to memory limits, the number of parallel environments (and the batch size) has been reduced from 500 to 50 for the 10,000-agent case and to 5 for the 100,000-agent case.

Across the tested scales, GradMAP consistently achieves the lowest operating cost and keeps the mean and worst-case values of daily maximum voltage violations and daily end-state violations among the lowest. Regarding the training time to convergence, GradMAP is up to 5.3$\times$ faster than GradMA, 13.4$\times$ faster than IPPO, and 133$\times$ faster than MAPPO. Training takes under 15 minutes for 1{,}000 agents and around 40 minutes for 100{,}000 agents. Relative to the Naive baseline, the mean daily maximum voltage violation decreases by 93.2\% and 96.4\% in the 1,000-agent and the 100,000-agent cases, respectively. The corresponding reductions in the worst-case daily maximum voltage violation are 75.7\% and 66.7\%. These results demonstrate the scalability and efficiency of the proposed GradMAP.

\begin{figure}[tb]
    \centering
    \includegraphics[width=\columnwidth]{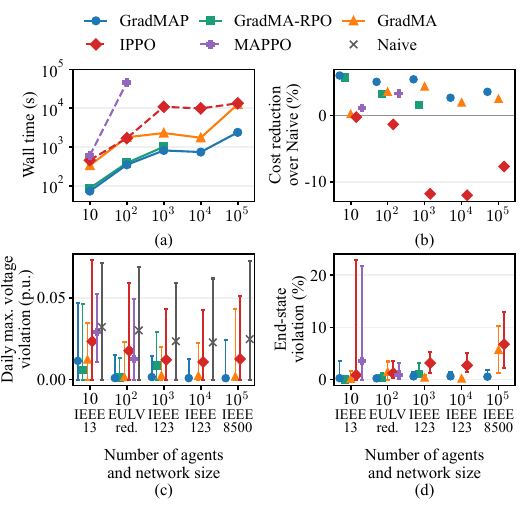}
    \vspace{-10mm}
    \caption{{\color{black}Scaling comparison across five agent population sizes and their corresponding distribution networks. (a) Wall-clock training time. (b) Cost reduction relative to the Naive baseline from 31-day totals; higher value indicates lower cost. (c,d) Averages (the central dots) and min--max ranges (the error bars) of 31 daily constraint-violation values. Each reported voltage violation is the daily maximum over buses, phases and time steps. The naive strategy has no end-state violations by construction.}}
    \label{fig:scale_trends}
\end{figure}
}

{\color{black}
\section{Discussion}\label{sec:discussion}

The proposed method requires environment gradients, which implies simulator-based learning. Theoretically, this scope is more restricted than that of standard RL approaches that can learn directly from the non-differentiable real world. However, it should be noted that power system applications are unlikely to accept learning directly from scratch in the real grid due to reliability and safety requirements. Simulator-based learning and validation is therefore a necessary process, in which the power grid's well-established model provides the required gradient information for the proposed GradMAP.

The proposed GradMAP follows the CTDE regime. Therefore, although we focus on simulator-based learning, it still requires a single entity to gather all the network and device information. However, this centralised information collection only takes place offline, which is a less restrictive setting than the online centralised regime that requires massive communication setup. A real-world example can be an aggregator which trains AI control policies for the grid-edge devices which it is responsible for coordinating, and the aggregator contacts with the network operator to obtain the network information needed to maintain network safety (UK network operators are making network parameters at 33~kV and above publicly available; access to network information can also be negotiated, because it would otherwise be difficult for the aggregator to ensure network safety).

Such simulator-based learning can introduce a sim-to-real gap, but in the adaptation stage, standard RL can be applied to adapt a pre-trained policy to the real-world grid \cite{deng2026supervised}. Our method is valuable for providing an efficient and effective pre-training strategy.
}

\vspace{-0mm}
\section{Conclusion}\label{sec:conclusion}
\vspace{-0mm}

This paper proposes GradMAP, a gradient-based multi-agent proximal learning framework for coordinating large-scale grid-edge flexibility in a fully decentralised manner. By exploiting the differentiability of power system environments and reusing exact gradients multiple times within a policy-output-space trust region (rather than a PPO-style distribution-space trust region), GradMAP achieves both computational efficiency and effective learning. 
\textcolor{black}{Using a single workstation-class NVIDIA RTX PRO 5000 Blackwell GPU with 48~GB of memory, GradMAP trains policies for 1{,}000 agents in under 15 minutes and 100{,}000 agents in around 40 minutes, without parameter sharing. In both cases, GradMAP achieve the lowest mean daily maximum voltage violation (more than 90\% below the Naive baseline) and the lowest energy cost among all evaluated methods, including standard MARL methods and gradient-based learning methods extended to the multi-agent setting by us. The training time to convergence is also the lowest, with a 5$\times$ to more than $10\times$ training speedup compared to other methods.}

Several directions remain for future work. One is to go beyond first-order gradients and incorporate higher-order derivative reuse into GradMAP, as commonly exploited in optimisation-based methods. Another is to extend GradMAP to discrete-action settings. One direct approach is to use the Straight-Through Estimator, which is biased but performs well empirically \cite{bengio2013estimating}. \textcolor{black}{Also, to address a fundamental challenge faced by all learning-based approaches, it is important to evaluate and improve the performance of GradMAP under topology perturbations, parameter uncertainty, device dropout, and unseen operating conditions.}
Finally, the proposed GradMAP method could also be promising for other decision-making domains where there are good differentiable environment simulators but evaluating gradients or higher-order derivatives is expensive, such as robotics and autonomous driving.

\appendix\label{app:ablation}

\begin{figure}[tb]
\centering
\includegraphics[width=\columnwidth]{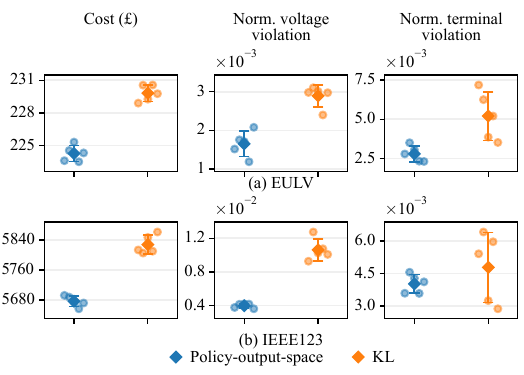}
\vspace{-8mm}
\caption{{\color{black}Regularisation using our policy-output-space squared distance versus PPO-style distribution-space KL penalties on (a) reduced EULV and (b) IEEE123, each with 100 agents. Columns show mean daily cost and normalised voltage/end-state violations of models trained using five different random seeds. Diamonds and whiskers mark their mean and standard deviation.}}
\label{fig:penalty-comparison}
\end{figure}

{\color{black}
Figure~\ref{fig:penalty-comparison} compares the proposed policy-output-space and KL penalties on the reduced EULV and IEEE123 (100 agents each, averaged over five seeds). The proposed policy-output-space regularisation penalty gives lower mean cost and constraint violations for these settings. 
}

{\color{black}
Table~\ref{tab:reuse-ablation} compares the output penalty and the number of gradient reuse proximal updates $K_{\text{prox}}$ on IEEE123 with 1,000 agents. Increasing reuse from 1 to 80 improves cost and voltage performance, but the benefits saturate when $K_{\text{prox}}$ is beyond 40 ($K_{\text{prox}}=80$ has lower cost and voltage violation but slightly higher device end-state constraint violation). Also, as can be seen in the first row, simple gradient reuse without regularisation leads to significant performance degradation, which highlights the importance of the proposed policy-output-space regularisation.
}

\begin{table}[tb]
\centering
\color{black}
\caption{Output-penalty and gradient-reuse ablations.}
\label{tab:reuse-ablation}
\setlength{\tabcolsep}{3pt}
\footnotesize
\begin{tabular}{@{}lrrr@{}}
\toprule
Variant & Cost (\pounds/day) & Voltage & End-state \\
\midrule
Without regularisation, $K_{\text{prox}}=80$ & 6{,}774.32 & 0.08620 & 0.05040 \\
$K_{\text{prox}}=1$ & 6{,}047.71 & 0.02063 & 0.11076 \\
$K_{\text{prox}}=10$ & 5{,}637.54 & 0.01437 & 0.02745 \\
$K_{\text{prox}}=40$ & 5{,}541.01 & 0.00640 & 0.00451 \\
$K_{\text{prox}}=80$ & 5{,}436.91 & 0.00583 & 0.00626 \\
\bottomrule
\end{tabular}
\end{table}

\vspace{-1mm}
\bibliographystyle{IEEEtran}
\begingroup
\let\originalbibitem\bibitem
\expandafter\def\csname revisedbib@zhong2025reparameterization\endcsname{}
\expandafter\def\csname revisedbib@deng2026supervised\endcsname{}
\expandafter\def\csname revisedbib@eltved2020robustness\endcsname{}
\renewcommand{\bibitem}[1]{%
    \color{black}%
    \ifcsname revisedbib@#1\endcsname\color{black}\fi
    \originalbibitem{#1}}
\bibliography{ref,response_refs}
\endgroup
\end{document}